\documentclass[sigconf]{acmart}

\AtBeginDocument{%
  }

\usepackage{booktabs}
\usepackage{multirow}
\usepackage[inline]{enumitem}
\usepackage{spverbatim}
\usepackage{tabularx}
\usepackage{fancyvrb}
\usepackage{alltt}
\usepackage{tcolorbox}
\usepackage{amsmath}
\usepackage{colortbl,array,xcolor}
\usepackage{graphicx}
\usepackage{arydshln}
\usepackage{afterpage}
\usepackage{comment}
\usepackage{subcaption}

\def\ojoin{\setbox0=\hbox{$\bowtie$}%
  \rule[-.02ex]{.25em}{.4pt}\llap{\rule[\ht0]{.25em}{.4pt}}}
\def\join{\mathbin{\mkern1.8mu\bowtie}}
\def\leftouterjoin{\mathbin{\ojoin\mkern-5.5mu\bowtie}}

\newcommand{\textsql}[1]{\texttt{#1}}
\newcommand{\textent}[1]{\textit{#1}}

\usepackage{bbding} 
\usepackage{soul}
\newcommand{\wrong}[1]{{\textbf{\color{red}$\times$#1}}} 
\newcommand{\correct}[1]{{\textbf{\color{blue}$\checkmark$#1}}} 
\newcommand{\highlight}[1]{\textbf{\ul{#1}}} 

\renewcommand\footnotetextcopyrightpermission[1]{} 

\begin{document}

\title{Exploring Database Normalization Effects {on} SQL Generation}

\author{Ryosuke Kohita}
\email{kohita\_ryosuke@cyberagent.co.jp}
\orcid{0009-0001-8414-9667}
\affiliation{%
  \institution{CyberAgent}
  \city{Tokyo}
  \country{Japan}
}


\begin{abstract}
Schema design, particularly normalization, is a critical yet often overlooked factor in natural language to SQL (NL2SQL) systems. 
Most prior research evaluates models on fixed schemas, overlooking the influence of design on performance.
We present the first systematic study of schema normalization's impact, evaluating eight leading large language models on synthetic and real-world datasets with varied normalization levels. 
We construct controlled synthetic datasets with formal normalization (1NF–3NF) and real academic paper datasets with practical schemes.
Our results show that denormalized schemas offer high accuracy on simple retrieval queries, even with cost-effective models in zero-shot settings. 
In contrast, normalized schemas (2NF/3NF) introduce challenges such as errors in base table selection and join type prediction; however, these issues are substantially mitigated by providing few-shot examples. 
For aggregation queries, normalized schemas yielded better performance, mainly due to their robustness against the data duplication and NULL value issues that cause errors in denormalized schemas.
These findings suggest that the optimal schema design for NL2SQL applications depends on the types of queries to be supported. Our study demonstrates the importance of considering schema design when developing NL2SQL interfaces and integrating adaptive schema selection for real-world scenarios.

\end{abstract}

\maketitle
\thispagestyle{plain} 
\pagestyle{plain}     

\section{Introduction}
Natural language to SQL (NL2SQL) systems have become increasingly important as they facilitate seamless translation between human intent and database operations, improving development processes, enhancing data analysis, and democratizing data access for non-technical users~\cite{yuankai-2023-gensql, antonis-2024-guiged}.
Recent advances in this field have been driven by improvements in architectures, algorithms, and benchmark datasets, especially with the emergence of large language models (LLMs)~\cite{gao2024dail}.
However, the impact of database-side factors such as schema design on NL2SQL performance has received limited attention.
It is known that complex database schemas make it more difficult for humans to write queries~\cite{borthick-2001-the-effects, bowen-2002-further-evidence}, and recent studies have suggested that similar schema complexity influences the accuracy of NL2SQL systems~\cite{ganti-2024-evaluating-real, luoma-2025-snails, zhang-2023-sciencebenchmark}.
By systematically investigating the influence of schema design on NL2SQL performance, we aim to provide new insights into how databases should be structured to better support these systems.

\begin{figure}[t]
    \centering
    \includegraphics[width=1.0\linewidth]{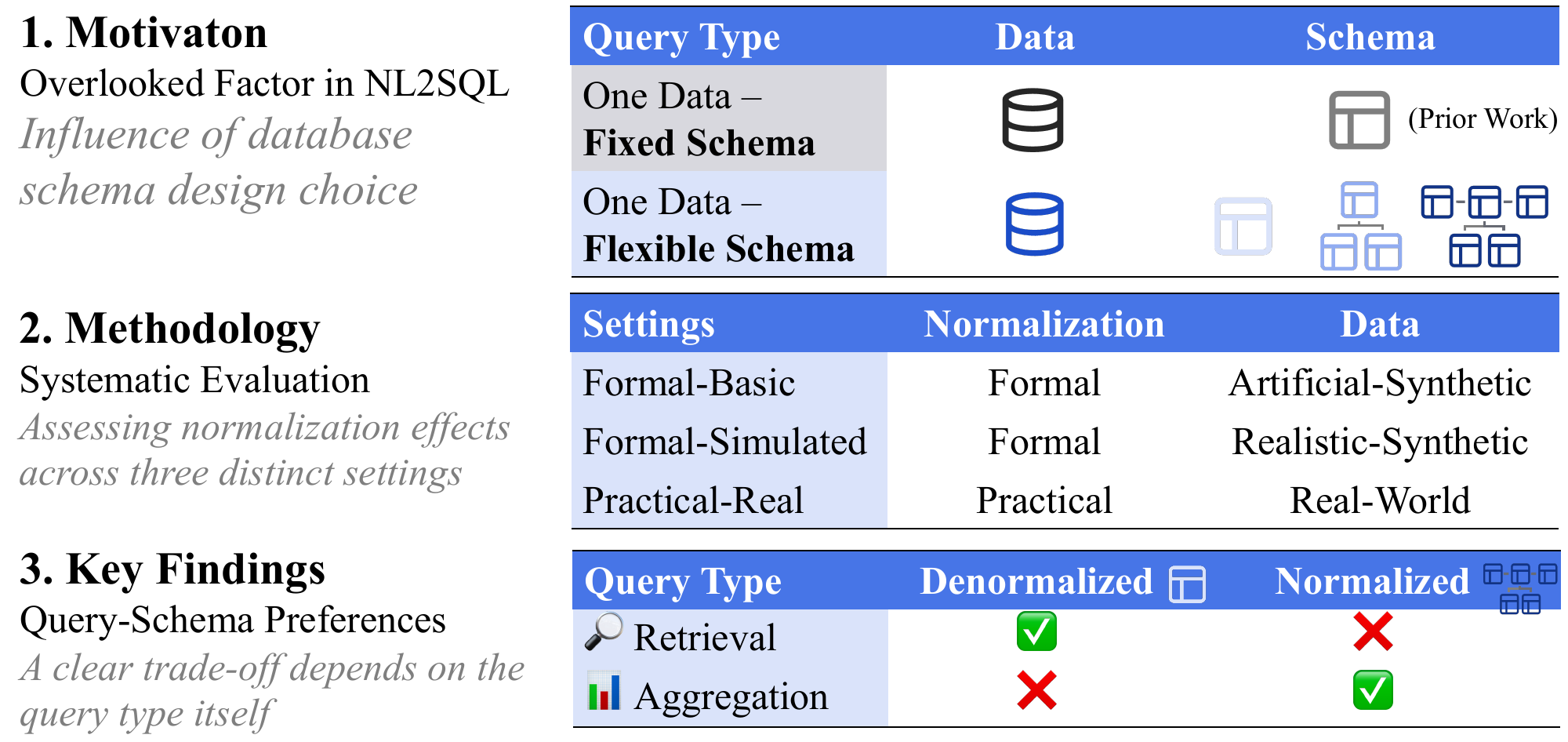}
    \caption{\small Overview of our motivation, methodology, and key finding—query-schema preferences. These preferences were consistently observed across experiments on leading eight LLMs.}
    \label{figure:overall}
\end{figure}

Research in NL2SQL has focused on model development. Various architectures and techniques have been proposed~\cite{liu2024surveynl2sqllargelanguage} such as decoding workflow~\cite{pourreza2024din, qu-etal-2024-generation, jiang-etal-2023-structgpt, fan-2024-combining}, fine-tuning~\cite{liu-2022-semantic, pourreza-rafiei-2024-dts, li-2024-codes, xu-etal-2024-symbol}, and task-oriented prompting~\cite{zihui-2023-few-shot-text, zhang-etal-2023-act, lee-etal-2025-mcs}. Also, efforts have been made to create larger and practical validation databases for benchmarks~\cite{zhongSeq2SQL2017, yu-etal-2018-spider, lee-2021-kaggle-dbqa, jinyang2024canllm}.
Nevertheless, most existing research and evaluation assumes a fixed, canonical database schema, even though in practice the same data is often structured in different ways to match real-world requirements. This reveals a gap between experimental validation and practical operation regarding the application of database design principles.

A central principle in database design is \emph{normalization}, a process of organizing data to minimize redundancy and improve data integrity~\cite{codd1970relational, codd1982relational}. While crucial for maintaining data consistency, it often increases schema complexity by creating more interconnected tables. This complexity, in turn, can make NL2SQL query generation more challenging. On the other hand, denormalization can simplify some queries but risks integrity issues~\cite{sanders2001denormalization, shin2006denormalization}. This inherent trade-off makes normalization a natural starting point for exploring how schema design influences NL2SQL systems.

To systematically investigate the impact of normalization, we designed a series of progressive experiments that move from a controlled, synthetic environment to a complex, real-world scenario, as illustrated in Figure~\ref{figure:overall}.
First, the \textsc{Formal-Basic} setting uses a minimal schema with formal normalization levels (1NF, 2NF, 3NF) to isolate the baseline effects of normalization on retrieval queries.
Second, \textsc{Formal-Simulated} introduces realistic domain contexts (flight scheduling, etc.) while maintaining experimental control.
Finally, \textsc{Practical-Real} validates our findings on actual academic publication data using three practical normalization approaches (LOW, MID, HIGH) that reflect design choices commonly seen in real-world systems, covering both retrieval and aggregation queries.

Our findings reveal distinct effects of normalization across experimental settings. In controlled environments (\textsc{Formal-Basic} and \textsc{Formal-Simulated}), denormalized single-table schemas consistently yield the highest NL2SQL performance, even for cost-effective models in zero-shot settings. Although increased normalization introduces challenges with table relationships, these can be mitigated by providing few-shot examples. In contrast, in the \textsc{Practical-Real} setting, normalized schemas generally outperform denormalized designs, particularly for aggregation queries where they better handle duplication and NULL values. For retrieval queries, the difference is less pronounced, and denormalized schemas remain competitive, especially when using cost-effective models. These results highlight that NL2SQL performance depends on both query types and normalization levels, underscoring the need to align schemas with actual workloads and model capabilities.

Our work offers important contributions to both research and practice. We provide the first comprehensive and systematic examination of how database normalization influences NL2SQL systems. Through experiments with eight production-grade LLMs, we derive practical insights: denormalization is effective when retrieval queries are dominant and a flat table is feasible, while moderate normalization is preferable for analytical workloads involving complex relationships or aggregation. We identify common error patterns, such as incorrect joins and duplicate handling, which suggest improving view design and model robustness.
Future NL2SQL systems may benefit from dynamically selecting or adapting schema variants based on query and model characteristics. Our work lays the foundation for this line of research and opens promising directions for developing more adaptable and effective NL2SQL solutions.

\section{Background}
\begin{figure}[t]
    \centering
    \includegraphics[width=0.975\linewidth]{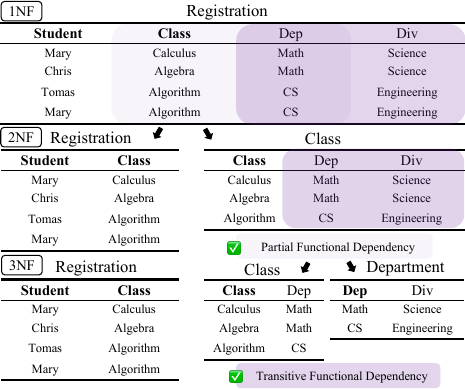}
    \caption{\small Normalization examples in 1NF, 2NF, and 3NF: class registrations with entities Student, Class, Department (Dep), and Division (Div). Functional dependencies: $\text{Class} \rightarrow \text{Dep}$ and $\text{Dep} \rightarrow \text{Div}$. Bold columns are primary keys, and the others are non-key attributes.}
    \label{figure:1nf-2nf-3nf-examples}
\end{figure}

Normalization is a database design principle that minimizes redundancy and prevents update anomalies by decomposing data into smaller tables where each fact is stored once~\cite{codd1970relational, date2012database}.

Figure~\ref{figure:1nf-2nf-3nf-examples} illustrates this decomposition process. A single denormalized table (1NF) contains redundant information (e.g., repeating the Science division for every Math department course). While this structure allows for simple \textsql{SELECT} statements, it is prone to data anomalies. Normalization resolves this by creating multiple, smaller tables (2NF, 3NF), ensuring each fact is stored only once.

However, this decomposition introduces a critical challenge for NL2SQL systems: retrieving information that spans these tables (e.g., finding the Division for a given Class) now requires generating queries with multiple \textsql{JOIN} operations. This complexity increases the risk of errors in table selection, join path inference, and overall query construction. Denormalization, conversely, simplifies queries by pre-joining tables, trading data consistency for performance~\cite{shin2006denormalization}. Our study focuses on the spectrum from 1NF to 3NF, as these levels are most relevant to practical applications~\cite{sebastian2016relational}.

\begin{table*}[t]
\centering
\small
\caption{\small Summary of experimental scenarios. Each row describes a schema/domain setting, the core entity triplet (FDT) used, a representative natural language query, and relationship patterns present. Scenarios and FDTs indicate the underlying real-world domain and entity structures. Abbreviations: 1:M=one-to-many, M:M=many-to-many, 1:1=one-to-one, MR=multi-role, Ret=retrieval, and Agg=aggregation.}
\label{tab:scenario-summary}
\begin{tabular}{llllcc}
\toprule
Schema-Data & Scenario & Query Examples & FDT / \#Tables & 1:M/M:M/1:1/MR/Ret/Agg \\
\toprule
\textsc{Formal-Basic}
  & basic
  & \footnotesize{\textit{List records where $a \geq 3$.}}
  & (\textent{A}, \textent{B}, \textent{C}) / 3
  & $\checkmark$/$\times$/$\times$/$\times/\checkmark$/$\times$
\\ 
\multirow{6}{*}{\textsc{Formal-Simulated}}
  & \multirow{2}{*}{flight}
  & \multirow{2}{*}{\footnotesize{\textit{List the flight schedules for the gate G0.}}}
  & (\textent{Flight}, \textent{Gate}, \textent{Terminal}) / 6
  & \multirow{2}{*}{$\checkmark$/$\times$/$\times$/$\times/\checkmark$/$\times$}
\\
  & & & (\textent{Flight}, \textent{Pilot}, \textent{License}) & &
\\
  & \multirow{2}{*}{library}
  & \multirow{2}{*}{\footnotesize{\textit{List the borrowing history of books titled Sun.}}}
  & (\textent{Book}, \textent{Title}, \textent{Author}) / 7
  & \multirow{2}{*}{$\checkmark$/$\times$/$\times$/$\checkmark/\checkmark$/$\times$}
\\
  & & & (\textent{Borrow}, \textent{Return}, \textent{Desk}) & &
\\
  & \multirow{2}{*}{class}
  & \multirow{2}{*}{\footnotesize{\textit{List the registration statuses for classes in the field Biology.}}}
  & (\textent{Student}, \textent{Professor}, \textent{Lab}) / 5
  & \multirow{2}{*}{$\checkmark$/$\checkmark$/$\times$/$\checkmark/\checkmark$/$\times$}
\\
  & & & (\textent{Class}, \textent{Professor}, \textent{Lab}) & &
\\
\multirow{2}{*}{\textsc{Practical-Real}}
   & \multirow{2}{*}{real}
   & \footnotesize{\textit{List papers by author 20343.}}
   & \multirow{2}{*}{- / 15}
   & \multirow{2}{*}{$\checkmark$/$\checkmark$/$\checkmark$/$\checkmark$/$\checkmark$/$\checkmark$}
\\
   & & \footnotesize{\textit{Count the number of papers in 2023 for each category.}} & & & 
\\
\bottomrule
\end{tabular}
\end{table*}

\section{Methodology}
\label{sec:methodology}
To systematically investigate how normalization influences SQL generation, we design three experimental settings that progress from a controlled, abstract environment to a complex, real-world scenario. This progressive approach allows us to first isolate the fundamental effects of normalization and then validate our findings under more realistic conditions.

\begin{enumerate}[label=(\arabic*)]
\item \textbf{\textsc{Formal-Basic} (\textsc{F-Basic})}: This initial experiment uses a minimal, fully artificial schema and data to establish a baseline understanding of how formal normalization levels (1NF, 2NF, 3NF) impact simple retrieval queries.
\item \textbf{\textsc{Formal-Simulated} (\textsc{F-Sim})}: Building on \textsc{F-Basic}, this setting introduces realistic domain contexts (e.g., flight management, library lending) to synthetic data, allowing us to assess performance in scenarios that mimic real-world entity relationships while maintaining experimental control.
\item \textbf{\textsc{Practical-Real} (\textsc{P-Real})}: Finally, this experiment employs a real-world academic papers dataset with three practical schema variants (LOW, MID, HIGH normalization). It covers both retrieval and aggregation queries to validate our findings in a scenario reflective of operational database design trade-offs.
\end{enumerate}

\subsection{Formal Normalization with Synthetic Data}
\label{sec:formal-normalization-with-synthetic-data}
We propose two synthetic experiments, \textsc{F-Basic} and \textsc{F-Sim}, using schemas designed via formal normalization principles. 
To ensure a controlled and comparable analysis, our methodology is built around the \textbf{Functional Dependency Triplet (FDT)}: a set of three entities $(A, B, C)$ linked by a transitive functional dependency $A \rightarrow B \rightarrow C$. This represents the minimal structure required to systematically study the decomposition process from 1NF to 3NF.

\paragraph{Schema}
\label{sec:basic-simulated-schema}
\begin{table}[t]
    \centering
    \small
    \caption{\small Schema layouts for each normalization level for an FDT $(A, B, C)$. 1NF: No join required (single table). 2NF: One join required between \texttt{A} and \texttt{B}. 3NF: Two joins required to connect \texttt{A}, \texttt{B}, and \texttt{C}.}
    \label{tab:schema-overview-basic}
    \begin{tabular}{lllllllll}
        \toprule
         \multirow{2}{*}{Schema / Table} & \multicolumn{8}{c}{Entities and Attributes} \\
        & \multicolumn{3}{c}{\textent{A}} & \multicolumn{3}{c}{\textent{B}} & \multicolumn{2}{c}{\textent{C}} \\
        \midrule
        1NF / \textsql{A}& \multicolumn{1}{>{\columncolor{gray!15}}l}{} & \multicolumn{3}{>{\columncolor{gray!15}}l}{a} & \multicolumn{1}{>{\columncolor{gray!15}}l}{b} & \multicolumn{2}{>{\columncolor{gray!15}}l}{} & \multicolumn{1}{>{\columncolor{gray!15}}l}{c} \\
        2NF / \textsql{A} $\mid$ \textsql{B} &  \multicolumn{1}{>{\columncolor{gray!15}}l}{id} & \multicolumn{1}{>{\columncolor{gray!15}}l}{a} & \multicolumn{1}{>{\columncolor{gray!15}}l}{$\text{B}_{\text{id}}$} & \multicolumn{1}{|>{\columncolor{gray!10}}l}{id} & \multicolumn{1}{>{\columncolor{gray!10}}l}{b} & \multicolumn{2}{>{\columncolor{gray!10}}l}{} & \multicolumn{1}{>{\columncolor{gray!10}}l}{c} \\
        3NF / \textsql{A} $\mid$ \textsql{B} $\mid$ \textsql{C} & \multicolumn{1}{>{\columncolor{gray!15}}l}{id} & \multicolumn{1}{>{\columncolor{gray!15}}l}{a} & \multicolumn{1}{>{\columncolor{gray!15}}l}{$\text{B}_{\text{id}}$} & \multicolumn{1}{|>{\columncolor{gray!10}}l}{id} & \multicolumn{1}{>{\columncolor{gray!10}}l}{b} & \multicolumn{1}{>{\columncolor{gray!10}}l}{$\text{C}_{\text{id}}$} & \multicolumn{1}{|>{\columncolor{gray!5}}l}{id} & \multicolumn{1}{>{\columncolor{gray!5}}l}{c} \\
        \bottomrule
    \end{tabular}     
\end{table}

Based on the FDT, we define three schema variants as illustrated in Table~\ref{tab:schema-overview-basic}. In \textbf{1NF}, all entities reside in a single, denormalized table, \textsql{A}. This is decomposed for \textbf{2NF}, where the partial dependency ($A \rightarrow B$) is resolved by splitting the data into two tables, \textsql{A} and \textsql{B}. Finally, in \textbf{3NF}, the transitive dependency ($B \rightarrow C$) is resolved by further splitting into a third table, \textsql{C}, resulting in a fully normalized, three-table schema.

\paragraph{Query}

\begin{table}
\centering
\small
\caption{\small Summary of query types, join operations, and outputs (for 3NF). Each type covers a subset of FDT entities $(A, B, C)$; the same query requires more joins as normalization increases; for example, query $A$ is a simple \textsql{SELECT * FROM A} in 1NF but requires two \textsql{LEFT JOIN}s as \textsql{SELECT * FROM A LEFT JOIN B AND LEFT JOIN C} in 3NF.}
\label{tab:join-patterns}
\begin{tabular}{lcc}
\toprule
Query Type & Join Operations & Output \\
\midrule
ABC & $\textsql{A} \join \textsql{B} \bowtie \textsql{C}$& $(A, B, C)$ \\
AB & $\textsql{A} \join \textsql{B} \leftouterjoin \textsql{C}$& $(A, B, C_\emptyset)$ \\
BC & $\textsql{B} \join \textsql{C} \leftouterjoin \textsql{A}$ & $(A_\emptyset, B, C)$ \\
A & $\textsql{A} \leftouterjoin \textsql{B} \leftouterjoin \textsql{C}$ & $(A, B_\emptyset, C_\emptyset)$ \\
B & $\textsql{B} \leftouterjoin \textsql{A} \leftouterjoin \textsql{C}$ & $(A_\emptyset, B, C_\emptyset)$ \\
C & $\textsql{C} \leftouterjoin \textsql{A} \leftouterjoin \textsql{B}$ & $(A_\emptyset, B_\emptyset, C)$ \\
\bottomrule
\end{tabular}
\parbox{\linewidth}{\footnotesize%
\vspace{2pt} 

$\join$ and $\leftouterjoin$ denote \textsql{INNER JOIN} and \textsql{LEFT OUTER JOIN}, respectively; $(A_\emptyset, ...)$ indicates possible \textsql{NULL}s for missing entities. AC is omitted as it is equivalent to ABC.}
\end{table}

We design six query types systematically covering retrieval patterns over FDT entities, as shown in Table~\ref{tab:join-patterns}. Each query type represents a specific retrieval pattern requiring different \textsql{JOIN} operations. \textsql{INNER JOIN (JOIN)} returns records only when a match exists in both tables, whereas \textsql{LEFT OUTER JOIN (LEFT JOIN)} returns all rows from the left table, inserting \textsql{NULL}s for non-matching rows. Consequently, as normalization increases, retrieving the same information requires more complex queries. For example, a query for entity A is a simple \textsql{SELECT} in 1NF but requires two \textsql{LEFT JOIN}s in 3NF to explicitly handle potential \textsql{NULL} values.

\paragraph{Data}
The two synthetic settings differ in their data and query formulation. In \textbf{\textsc{F-Basic}}, we use a minimal FDT with simple integer attributes, and natural language questions closely mirror the SQL structure (e.g., \textit{List records where $a \geq 3$}) to isolate the impact of schema complexity. In contrast, \textbf{\textsc{F-Sim}} introduces three realistic scenarios (flight scheduling, library lending, class registration) with more complex FDTs (e.g., (\textent{Flight}, \textent{Gate}, \textent{Terminal})). As summarized in Table~\ref{tab:scenario-summary}, these scenarios include richer semantics such as many-to-many relationships (e.g., \textent{Student} and \textent{Class}) and multi-role entities (e.g., \textent{Professor} as advisor or instructor). Queries emulate natural user requests (e.g., \textit{List the flight schedules for gate G0}), providing a near-practical yet controlled evaluation environment.

\subsection{Practical Normalization with Actual Data}
\label{sec:practical-normalization-with-actual-data}
To complement our synthetic experiments, we evaluate performance on a real-world dataset of academic papers from Semantic Scholar~\cite{Kinney2023TheSS}. While formal normalization is theoretically rigorous, practical schema design often requires balancing such rigor with operational needs. Therefore, we created three schema variants reflecting common engineering trade-offs: \textbf{LOW}, \textbf{MID}, and \textbf{HIGH} normalization. These designs allow us to examine the impact of redundancy versus query complexity in a realistic setting.

\paragraph{Schema}
We created three schema variants reflecting common engineering trade-offs. The \textbf{HIGH} schema is a fully normalized schema for data consistency, where each entity and relationship is separated into a distinct table (e.g., separating author and citation statistics). The \textbf{MID} schema balances integrity with practical simplicity by, for example, embedding one-to-one attributes (like citation statistics) directly within the author table. Finally, the \textbf{LOW} schema is a denormalized schema optimized for frequent retrieval. While core entities (papers and authors) remain separate to avoid data explosion, supplementary fields are duplicated to reduce joins.

\paragraph{Query}
\label{section:real-world-experiment-question}
For this setting, we constructed a diverse set of 26 retrieval and 29 aggregation query templates to reflect realistic needs such as author lookups, citation analysis, and venue-based statistics. Initial templates were generated using GPT-4o, then manually selected and refined to ensure broad coverage of topics and complexity levels. All finalized queries were manually implemented for each of the three schema variants to create our ground truth.

\paragraph{Data}
\label{section:real-world-experiments-data}
The resulting dataset provides a realistic testbed for our experiments, featuring common real-world characteristics. Our data consists of papers from Semantic Scholar that mention ``large language model,'' along with their authors and cited papers. Crucially, the dataset includes complexities such as missing values, authors with multiple affiliations, and unmerged entity references represented by string names (e.g., ``Google'' vs. ``Google Research''). These features create a robust environment for examining the effects of schema normalization on NL2SQL performance.

\section{Experiment}
\subsection{Settings}
\label{sec:experimental-setup}
To evaluate the impact of schema normalization, we conducted experiments using eight production-grade LLMs from the GPT, Gemini, and Claude families.\footnote{
Models used: GPT-4o-mini (\texttt{gpt-4o-mini-2024-07-18}), GPT-4o (\texttt{gpt-4o-2024-08-06}), GPT-4.1-mini (\texttt{gpt-4.1-mini-2025-04-14}), GPT-4.1 (\texttt{gpt-4.1-2025-04-14}); Gemini 1.5 Pro (\texttt{gemini-1.5-pro}), Gemini 2.0 Flash (\texttt{gemini-2.0-flash}); Claude 3.5 Sonnet (\texttt{claude-3-5-sonnet-20241022}), Claude 3.7 Sonnet (\texttt{claude-3-7-sonnet-20250219}).
All schema definitions, prompts, queries, and data are available at \url{https://github.com/CyberAgentAILab/exploring-dbnorm}
}
For the \textsc{F-Basic} and \textsc{F-Sim} experiments, we generated three datasets per scenario using scenario-specific probability models for
realism. Each schema involved six canonical query types ($\S$~\ref{sec:formal-normalization-with-synthetic-data}). For the \textsc{P-Real} experiment, we used our Semantic Scholar dataset with 55 diverse query templates (26 retrieval and 29 aggregation) ($\S$~\ref{sec:practical-normalization-with-actual-data}). In all experiments, each template was instantiated with five different filter conditions.

We evaluated performance using standard execution accuracy~\cite{yu-etal-2018-spider}, and queries with a computation time exceeding 60 seconds were marked incorrect. Both zero-shot and few-shot settings were tested. In the latter, five demonstration pairs (natural language request and corresponding SQL) were provided as in-context examples. All experiments used a minimal, standardized setup to focus on normalization effects under practical, out-of-the-box conditions, leaving advanced workflow optimizations for future work.

\subsection{Results on Synthetic Data (\textsc{F-Basic} \& \textsc{F-Sim})}

\subsubsection{Performance Trends}
\begin{table}[t]
\centering
\small
\caption{\small Execution Accuracy in \textsc{Formal-Basic} (95\% CI).}
\label{tab:scores-basic}
\begin{tabular}{lllll}
\toprule
 Fewshot   & Model             & 1NF          & 2NF          & 3NF          \\
\midrule
 \multirow{8}{*}{zero} & GPT-4o-mini       & 1.00 (±0.00) & 0.38 (±0.10) & 0.21 (±0.08) \\
 & GPT-4o            & 1.00 (±0.00) & 0.38 (±0.10) & 0.21 (±0.08) \\
 & GPT-4.1-mini      & 1.00 (±0.00) & 0.51 (±0.10) & 0.26 (±0.09) \\
 & GPT-4.1           & 1.00 (±0.00) & 0.38 (±0.10) & 0.21 (±0.08) \\
 & Gemini 1.5    & 1.00 (±0.00) & 0.38 (±0.10) & 0.21 (±0.08) \\
 & Gemini 2.0  & 1.00 (±0.00) & 0.38 (±0.10) & 0.21 (±0.08) \\
 & Claude 3.5 & 1.00 (±0.00) & 0.51 (±0.10) & 0.21 (±0.08) \\
 & Claude 3.7 & 1.00 (±0.00) & 0.46 (±0.10) & 0.22 (±0.09) \\
 \midrule
 \multirow{8}{*}{few} & GPT-4o-mini       & 1.00 (±0.00) & 0.64 (±0.10) & 0.59 (±0.10) \\
 & GPT-4o            & 1.00 (±0.00) & 0.92 (±0.06) & 0.91 (±0.06) \\
 & GPT-4.1-mini      & 1.00 (±0.00) & 0.81 (±0.08) & 0.80 (±0.08) \\
 & GPT-4.1           & 1.00 (±0.00) & 0.82 (±0.08) & 0.89 (±0.07) \\
 & Gemini 1.5    & 1.00 (±0.00) & 0.93 (±0.05) & 0.91 (±0.06) \\
 & Gemini 2.0  & 1.00 (±0.00) & 0.81 (±0.08) & 0.94 (±0.05) \\
 & Claude 3.5 & 1.00 (±0.00) & 0.94 (±0.05) & 0.92 (±0.06) \\
 & Claude 3.7 & 1.00 (±0.00) & 0.86 (±0.07) & 0.92 (±0.06) \\
\bottomrule
\end{tabular}
\end{table}

\begin{table}[t]
    \centering
    \small
    \caption{\small Execution accuracy in \textsc{Formal-Simulated} (95\% CI).}
    \label{tab:scores-simulated}
    \begin{tabular}{lllll}
        \toprule
         Fewshot   & Model             & 1NF          & 2NF          & 3NF          \\
        \midrule
         \multirow{8}{*}{zero} & GPT-4o-mini       & 0.87 (±0.03) & 0.47 (±0.04) & 0.30 (±0.04) \\
         & GPT-4o            & 0.99 (±0.01) & 0.71 (±0.04) & 0.65 (±0.04) \\
         & GPT-4.1-mini      & 0.99 (±0.01) & 0.69 (±0.04) & 0.60 (±0.04) \\
         & GPT-4.1           & 1.00 (±0.01) & 0.77 (±0.04) & 0.68 (±0.04) \\
         & Gemini 1.5 & 1.00 (±0.00) & 0.60 (±0.04) & 0.49 (±0.04) \\
         & Gemini 2.0 & 0.99 (±0.01) & 0.49 (±0.04) & 0.49 (±0.04) \\
         & Claude 3.5 & 0.99 (±0.01) & 0.73 (±0.04) & 0.63 (±0.04) \\
         & Claude 3.7 & 1.00 (±0.00) & 0.75 (±0.04) & 0.61 (±0.04) \\
         \midrule
         \multirow{8}{*}{few} & GPT-4o-mini       & 0.99 (±0.01) & 0.90 (±0.03) & 0.71 (±0.04) \\
         & GPT-4o            & 1.00 (±0.00) & 0.95 (±0.02) & 0.94 (±0.02) \\
         & GPT-4.1-mini      & 1.00 (±0.00) & 0.96 (±0.02) & 0.93 (±0.02) \\
         & GPT-4.1           & 1.00 (±0.01) & 0.97 (±0.01) & 0.96 (±0.02) \\
         & Gemini 1.5    & 1.00 (±0.00) & 0.98 (±0.01) & 0.96 (±0.02) \\
         & Gemini 2.0  & 1.00 (±0.00) & 0.94 (±0.02) & 0.93 (±0.02) \\
         & Claude 3.5 & 1.00 (±0.00) & 0.98 (±0.01) & 0.93 (±0.02) \\
         & Claude 3.7 & 1.00 (±0.00) & 0.96 (±0.02) & 0.93 (±0.02) \\
        \bottomrule
    \end{tabular}
\end{table}

Table~\ref{tab:scores-basic} shows the results in \textsc{F-Basic}. All models achieved perfect accuracy for flat schemas (1.0 in 1NF) in both zero-shot and few-shot settings. However, performance dropped with increasing normalization level: in 2NF and 3NF, zero-shot accuracy fell to below 0.5 or even 0.3 for most models. Few-shot prompting improved performance for 2NF and 3NF, but this improvement was limited for smaller models like GPT-4o-mini.

Table~\ref{tab:scores-simulated} reports the results for \textsc{F-Sim}, which introduced more realistic entities and relationships. The experiment mirrored the trend from \textsc{F-Basic}, where accuracy declined as normalization increased.
A key difference, however, was the significantly higher overall performance. In the zero-shot setting, for instance, several leading models maintained accuracies above 0.60 even on the most complex 3NF schema.
Furthermore, few-shot prompting yielded substantial improvements, increasing accuracy for most models to over 0.90 for both 2NF and 3NF schemas. Notably, GPT-4o-mini also showed a significant increase, from 0.30 to 0.71 in 3NF.

\subsubsection{Error Analysis}
\begin{figure*}[t]
    \centering
    \includegraphics[width=1.0\linewidth]{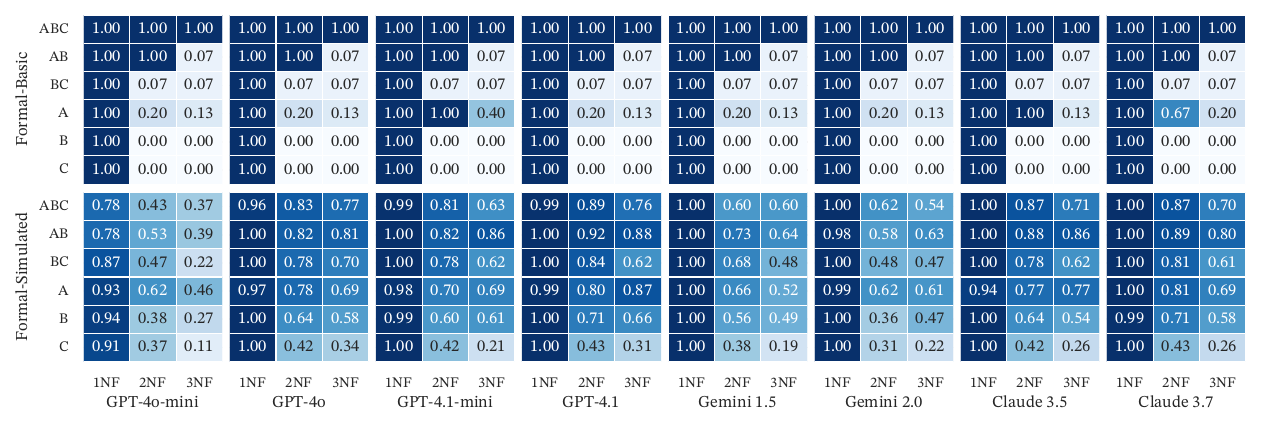}    
    \caption{\small Per-query-type execution accuracy for 1NF / 2NF / 3NF in zero-shot settings. Top: \textsc{Formal-Basic}; Bottom: \textsc{Formal-Simulated}.}
    \label{fig:combined-scores-query-type}
\end{figure*}

\label{sec:control-error-analysis}
\begin{table}[t]
\centering
\small
\caption{\small Error cases in the \textsc{Formal-Basic} and \textsc{Formal-Simulated} ($\times$ and $\checkmark$ denote wrong statements and their corrections).}
\label{tab:basic-simulated-error-cases}  
\begin{tabular}{p{8cm}}
\toprule
(a) Incorrect Join Type Selection \\ 
\begin{tcolorbox}[
    width=8.1cm,
    colback=gray!10,
    arc=3pt,
    boxrule=0pt,
    left=2pt,
    right=2pt,
    top=2pt,
    bottom=2pt,
    nobeforeafter
]
    \scriptsize
    \begin{Verbatim}[commandchars=\\\{\}]
SELECT ... FROM C \wrong{JOIN}(\correct{LEFT JOIN}) B ON C.id = B.C_id ...
    \end{Verbatim}
\end{tcolorbox} \\ \midrule

(b) Incorrect Base Table Selection \\ 
\begin{tcolorbox}[
    width=8.1cm,
    colback=gray!10,
    arc=3pt,
    boxrule=0pt,
    left=2pt,
    right=2pt,
    top=2pt,
    bottom=2pt,
    nobeforeafter
]
    \scriptsize
    \begin{Verbatim}[commandchars=\\\{\}]
SELECT ... FROM (\wrong{A}/\correct{B}) LEFT JOIN (\wrong{B}/\correct{A}) ON ... 
    \end{Verbatim}
\end{tcolorbox} \\ \midrule

(c) Table Confusion \\ 
\begin{tcolorbox}[
    width=8.1cm,
    colback=gray!10,
    arc=3pt,
    boxrule=0pt,
    left=2pt,
    right=2pt,
    top=2pt,
    bottom=2pt,
    nobeforeafter
]
    \scriptsize
    \begin{Verbatim}[commandchars=\\\{\}]
SELECT ... b.donated_by, (\wrong{b}/\correct{br}).due_date, ... FROM ...
    LEFT JOIN borrows \highlight{br} ON r.id = br.id
    LEFT JOIN books \highlight{b} ON br.book_id = b.id ...
\end{Verbatim}
\end{tcolorbox} \\

\begin{tcolorbox}[
    width=8.1cm,
    colback=gray!10,
    arc=3pt,
    boxrule=0pt,
    left=2pt,
    right=2pt,
    top=2pt,
    bottom=2pt,
    nobeforeafter
]
    \scriptsize
    \begin{Verbatim}[commandchars=\\\{\}]
SELECT ..., (\wrong{professor}/\correct{advisor}).name, (\wrong{professor}/\correct{advisor}).lab, ...
FROM registration
    LEFT JOIN professor AS \highlight{advisor} ON ...
    LEFT JOIN professor AS \highlight{instructor} ON ...
\end{Verbatim}
\end{tcolorbox} \\
\bottomrule
\end{tabular}
\end{table}

To understand the challenges normalization introduces, we analyzed the per-query-type results from the zero-shot setting (Figure~\ref{fig:combined-scores-query-type}) and found common error patterns (Table~\ref{tab:basic-simulated-error-cases}).

The performance drop in \textsc{F-Basic} was strongly correlated with the need for \textsql{LEFT JOIN} to handle potentially missing data (Figure~\ref{fig:combined-scores-query-type}, top). This challenge was evident in queries on the primary entity (A in 2NF/3NF) and partially-related entities (AB in 3NF), where models frequently used \textsql{JOIN} instead of \textsql{LEFT JOIN}, leading to the erroneous omission of rows (Table~\ref{tab:basic-simulated-error-cases}-a). In contrast, queries like AB in 2NF remained easy as they did not require \textsql{LEFT JOIN}.

This difficulty was compounded for queries targeting dependent entities (B, C, and BC). In addition to the \textsql{LEFT JOIN} challenge, these queries required the counter-intuitive step of setting a dependent entity as the query's base table. Models consistently failed at this, defaulting instead to entity A as the starting point (Table~\ref{tab:basic-simulated-error-cases}-b). This combination of errors explains the particularly severe performance drop observed for these query types.

The \textsc{F-Sim} setting introduced new semantic challenges where models would confuse the meaning of entities and roles, leading to semantic errors such as table confusion (Table~\ref{tab:basic-simulated-error-cases}-c). Although the fundamental join errors from \textsc{F-Basic} persisted, \textsc{F-Sim}'s overall accuracy was higher (Figure~\ref{fig:combined-scores-query-type}, bottom). We attribute this to two factors. First, the schema's natural semantics may have provided contextual clues that helped models avoid some join errors~\cite{luoma-2025-snails}. More significantly, the data's low rate of null values often masked the impact of incorrect joins, as a query using \textsql{JOIN} could still produce the correct result if no nulls needed to be preserved~\cite{zhong-etal-2020-semantic}.

Finally, we found that most of these error patterns could be effectively mitigated with few-shot prompting. By providing a handful of examples, models were able to learn the domain-specific rules for table management, such as the correct join paths and the appropriate use of \textsql{LEFT JOIN}s, greatly reducing these errors.

\subsection{Results on Real-World Data (\textsc{P-Real})}
\begin{table}[t]
\centering
\small
\caption{\small Execution accuracy in \textsc{Practical-Real} (95\% CI).}
\label{tab:scores-real}
\begin{tabular}{lllll}
\toprule
 Fewshot   & Model             & LOW          & MID          & HIGH         \\
\midrule
 \multirow{8}{*}{zero} & GPT-4o-mini       & 0.43 (±0.03) & \textbf{0.59} (±0.03) & 0.54 (±0.03) \\
 & GPT-4o            & 0.39 (±0.03) & \textbf{0.74} (±0.03) & \textbf{0.75} (±0.03) \\
 & GPT-4.1-mini      & 0.54 (±0.03) & \textbf{0.79} (±0.03) & 0.78 (±0.03) \\
 & GPT-4.1           & 0.61 (±0.03) & \textbf{0.79} (±0.03) & \textbf{0.79} (±0.03) \\
 & Gemini 1.5    & 0.20 (±0.03) & \textbf{0.60} (±0.03) & 0.55 (±0.03) \\
 & Gemini 2.0  & 0.33 (±0.03) & \textbf{0.64} (±0.03) & \textbf{0.63} (±0.03) \\
 & Claude 3.5 & 0.56 (±0.03) & \textbf{0.79} (±0.03) & 0.75 (±0.03) \\
 & Claude 3.7 & 0.55 (±0.03) & \textbf{0.81} (±0.03) & \textbf{0.79} (±0.03) \\
 \midrule

 \multirow{8}{*}{few} & GPT-4o-mini       & 0.45 (±0.03) & \textbf{0.59} (±0.03) & \textbf{0.56} (±0.03) \\
 & GPT-4o            & 0.52 (±0.03) & \textbf{0.76} (±0.03) & \textbf{0.77} (±0.03) \\
 & GPT-4.1-mini      & 0.59 (±0.03) & \textbf{0.80} (±0.03) & \textbf{0.77} (±0.03) \\
 & GPT-4.1           & 0.68 (±0.03) & \textbf{0.80} (±0.03) & \textbf{0.81} (±0.03) \\
 & Gemini 1.5    & 0.35 (±0.03) & \textbf{0.70} (±0.03) & 0.64 (±0.03) \\
 & Gemini 2.0  & 0.35 (±0.03) & 0.67 (±0.03) & \textbf{0.73} (±0.03) \\
 & Claude 3.5 & 0.66 (±0.03) & \textbf{0.82} (±0.03) & \textbf{0.81} (±0.03) \\
 & Claude 3.7 & 0.66 (±0.03) & \textbf{0.81} (±0.03) & \textbf{0.81} (±0.03) \\
\bottomrule
\end{tabular}
\end{table}

\begin{figure*}[t]
    \centering
    \includegraphics[width=1.0\linewidth]{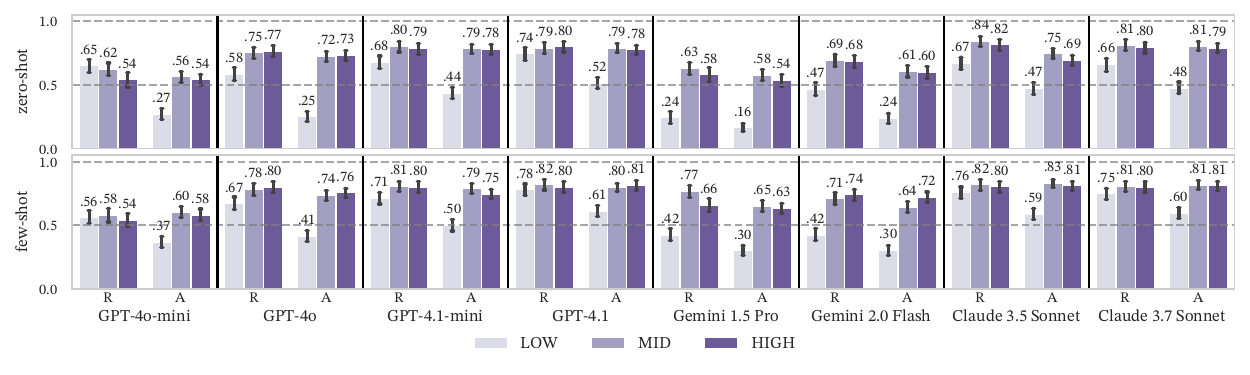}
    \caption{\small Execution accuracy for Retrieval (R) and Aggregation (A) queries at normalization levels (LOW, MID, HIGH) in \textsc{Practical-Real}.}
    \label{fig:real-scores-query-type}
\end{figure*}

\subsubsection{Performance Trends and Analysis by Query Type}
In the \textsc{P-Real} setting, we observed a notable reversal of the trend from our synthetic experiments: normalized schema designs held a clear advantage (Table~\ref{tab:scores-real}). Across all models, both the MID and HIGH schemas significantly outperformed the denormalized LOW schema. The distinction between MID and HIGH was minimal, with MID often showing slightly better performance. On these normalized schemas, leading models from the GPT-4.1 and Claude families achieved high scores of 0.79--0.82. It is also apparent from the table that few-shot prompting offered only marginal gains over the zero-shot setting. This limited impact is reflected in the average score improvements, which were approximately 0.08 for the LOW schema, and a mere 0.02--0.03 for the MID and HIGH schemas.

A breakdown of the results by query type provides insight into this trend (Figure~\ref{fig:real-scores-query-type}). For retrieval queries, the performance differences between schemas were generally small. The scores for the LOW schema were often comparable to those for MID and HIGH, with the performance gap averaging a modest 0.13 points. In contrast, for aggregation queries, the superiority of normalized schemas was much more pronounced. The LOW schema consistently underperformed in this case, resulting in scores that were, on average, 0.30 points lower than its normalized counterparts.

\subsubsection{Error Analysis}
\begin{table}[t]
\centering
\small
\caption{\small Error cases in the LOW schema of the \textsc{Practical-Real} experiment ($\times$ and $\checkmark$ denote wrong statements and their corrections).}
\label{tab:real-error-cases}
\begin{tabular}{p{8cm}}
\toprule
(a) Duplicate Record Selection \\ 
\begin{tcolorbox}[
    width=8.1cm,
    colback=gray!10,
    arc=3pt,
    boxrule=0pt,
    left=2pt,
    right=2pt,
    top=2pt,
    bottom=2pt,
    nobeforeafter
]
\scriptsize
    \begin{Verbatim}[commandchars=\\\{\}]
SELECT \correct{DISTINCT} P.id, P.title FROM papers P WHERE P.category =
    \end{Verbatim}
\end{tcolorbox}
\\ \midrule

(b) Missing Null Filtering \\ 
\begin{tcolorbox}[
    width=8.1cm,
    colback=gray!10,
    arc=3pt,
    boxrule=0pt,
    left=2pt,
    right=2pt,
    top=2pt,
    bottom=2pt,
    nobeforeafter
]
\scriptsize
    \begin{Verbatim}[commandchars=\\\{\}]
SELECT ..., COUNT(DISTINCT p.id) FROM papers p
    JOIN paper_authors pa ON ... JOIN authors a ON ...
WHERE p.year >= 2021
    \correct{AND p.publication_type IS NOT NULL AND a.affiliation IS NOT NULL}
GROUP BY a.affiliation, p.publication_type;
    \end{Verbatim}
\end{tcolorbox}
\\ \midrule

(c) Duplicate Record Counting \\ 
\begin{tcolorbox}[
    width=8.1cm,
    colback=gray!10,
    arc=3pt,
    boxrule=0pt,
    left=2pt,
    right=2pt,
    top=2pt,
    bottom=2pt,
    nobeforeafter
]
\scriptsize
    \begin{Verbatim}[commandchars=\\\{\}]
SELECT ..., COUNT(\wrong{*} / \correct{DISTINCT P.id}) FROM papers P
WHERE P.pdf_url IS NOT NULL AND P.year >= 2021 GROUP BY P.category;
    \end{Verbatim}
\end{tcolorbox}
\\
\bottomrule
\end{tabular}
\end{table}
The key to understanding the performance reversal in the \textsc{P-Real} setting lies in the distinctive error patterns of the denormalized LOW schema, particularly for aggregation queries. While the LOW schema simplified some join paths, its data redundancy introduced new critical errors. As illustrated in Table~\ref{tab:real-error-cases}, models consistently struggled to handle this redundancy, leading to: (a) duplicate records from omitting \textsql{DISTINCT}, (b) incorrect grouping due to missing null filters, and (c) significant overcounting from using \textsql{COUNT(*)} instead of \textsql{COUNT(DISTINCT ...)}.

These deduplication and null-handling challenges are addressed \emph{by design} in normalized schemas, which guarantee data integrity. In contrast, the denormalized LOW schema shifts this responsibility to the model, requiring it to generate complex, \emph{ad-hoc logic} at the query level. This task proved challenging for the models and was a key factor contributing to the LOW schema's underperformance.

Furthermore, the fundamental join-related errors observed in the synthetic experiments persisted across all three schema variants, as even the LOW schema required core joins. Unlike in the controlled synthetic setting, however, few-shot prompting offered little improvement. The diversity and complexity of real-world queries proved too great for models to generalize from a small set of examples, explaining the limited performance gains.

\subsection{Supplementary Analysis}
\begin{figure*}[!t]
  \centering
    \begin{minipage}[t]{0.48\linewidth}
    \vspace{0pt}
    \includegraphics[width=1.0\linewidth]{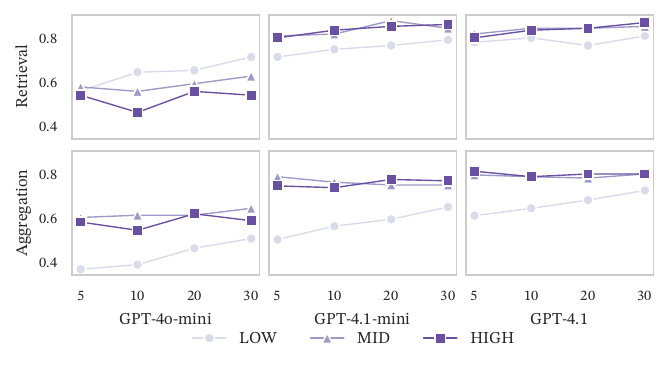}
  \end{minipage}
    \hfill
  \begin{minipage}[t]{0.48\linewidth}
    \vspace{0pt}
    \includegraphics[width=\linewidth]{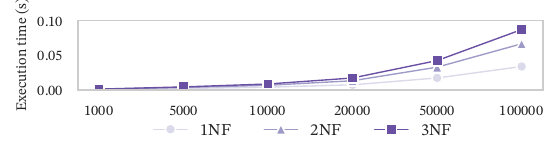}
    \vspace{8pt}
    \includegraphics[width=\linewidth]{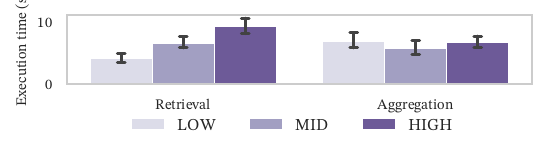}
  \end{minipage}
  \caption{\small (Left) Execution accuracy in \textsc{Practical-Real} with varying few-shot examples. (Right-Top) Execution times with different data volumes in \textsc{Formal-Basic}. (Right-Bottom) Execution times by query type in \textsc{Practical-Real}.}
  \label{fig:supplementary-analysis}
\end{figure*}
We also report on the influence of few-shot example count on model accuracy and briefly assess differences in execution speed among schema variants.

Figure~\ref{fig:supplementary-analysis} (left) shows how the number of few-shot examples affects accuracy. In the upper panels, execution accuracy for retrieval queries improved with more examples, particularly in the LOW schema. This effect was most pronounced for GPT-4o-mini, where the advantage of the LOW schema grew as the number of examples increased. In the lower panels, more examples also improved accuracy for aggregation queries on the LOW schema, but not for the normalized (MID, HIGH) schemas. This suggests that few-shot examples are particularly effective for denormalized schemas, helping models learn ad-hoc strategies. In contrast, for aggregation queries on normalized schemas, additional examples yielded little improvement. This indicates a potential limitation in learning complex, structural relationships from a small set of case-based examples.

Figure~\ref{fig:supplementary-analysis} (right) reports execution speed. As expected, the upper panel demonstrates that normalization increases execution time in \textsc{F-Basic}; the overhead of join operations means 3NF schemas incur higher computational costs than 1NF as the number of records increases. The lower panel illustrates the results for \textsc{P-Real}. Here, denormalized schemas led to faster retrieval queries, while for aggregation queries, timing differences were minimal, likely because the cost of the aggregation operation itself was the dominant factor.

\section{Discussion}
\label{sec:discussion}

\paragraph{Denormalization versus Normalization: A Query-Dependent Trade-off}
Our experiments showed that the optimal degree of normalization for NL2SQL systems is query-dependent. Denormalized (flat) schemas facilitate simple retrieval queries, often attaining high accuracy even with cost-effective models in zero-shot settings. In contrast, normalized schemas aid aggregation tasks, as they mitigate errors caused by data duplication and null handling. While few-shot examples could overcome some challenges of normalization in controlled synthetic settings, these errors often persisted in the more complex real-world scenario. This highlights the difficulty of designing systems that generalize across schema variants.

\paragraph{Implications for Practical NL2SQL Systems}
These findings have practical implications for NL2SQL systems, suggesting a shift from a single, static schema to a dynamic, workload-aware approach. An effective strategy is to maintain multiple schema variants tailored to different query types; for instance, creating denormalized materialized views for retrieval-heavy applications while using normalized base tables for analytical queries where data integrity is critical. This principle could be taken a step further by building advanced NL2SQL systems with an adaptive schema selection module. Such a module would classify a user's query intent (e.g., retrieval vs. aggregation) and route the SQL generation task to the most appropriate schema variant. This approach, inspired by operational best practices like query routing, represents a promising direction for creating more robust and accurate NL2SQL interfaces.

\paragraph{Limitations and Future Work}
Our study provides the first systematic analysis of normalization's effects, but it has several limitations for future research. First, our real-world experiments were limited to a single domain (academic publications), and future work should validate these findings across diverse domains and benchmarks for generalizability. Furthermore, our focus on \textsql{SELECT} queries leaves other database-side functionalities unexplored; investigating the impact on data modification operations (\textsql{INSERT}, \textsql{UPDATE}, \textsql{DELETE}) and other features like indexing or data types is valuable. Finally, a key future direction is to realize the vision outlined in our discussion: automating the generation and dynamic selection of schema variants to enable scalable, real-time adaptation in NL2SQL pipelines.

\section{Related Work}
NL2SQL research has rapidly advanced with the emergence of LLMs, which have enabled significant improvements in SQL generation across benchmarks \cite{gao2024dail, hong2024nextgenerationdatabaseinterfacessurvey}. A variety of architectural innovations, such as schema linking~\cite{li2023resdsql, dong2023c3zeroshottexttosqlchatgpt}, template selection~\cite{tonghui-2024-purple, gao2024dail, fu-2023-catsql}, and human collaboration~\cite{narechania-2021-diy, tian-etal-2023-interactive}, have contributed to this progress~\cite{liu2024surveynl2sqllargelanguage}. Dataset development has paralleled these advances, evolving from single-table settings (WikiSQL~\cite{zhongSeq2SQL2017}) to multi-table, cross-domain, and enterprise-scale resources such as Spider~\cite{yu-etal-2018-spider}, BIRD~\cite{jinyang2024canllm}, and Spider 2.0~\cite{lei2024spider20evaluatinglanguage}. These recent benchmarks highlight persistent challenges posed by complex, realistic database schemas~\cite{floratou-2024-nl2sql-is-a-solved, boyan-2024-dawn}.

Despite these advances, the role of database schema design remains underexplored in NL2SQL research. Early studies in database systems and human factors found that higher normalization increases query complexity and the risk of errors for users~\cite{borthick-2001-the-effects, bowen-2002-further-evidence}. Later work in the NL2SQL domain has observed similar issues, noting that model accuracy declines as schema complexity rises due to difficulties in understanding relationships~\cite{ganti-2024-evaluating-real, floratou-2024-nl2sql-is-a-solved, mitsopoulou-2025-analysis}. 
Recent proposals address schema-related challenges through techniques such as schema pruning, routing, and contextual prompts~\cite{shen-etal-2024-improving, wang-etal-2025-mac, talaei2024chess}. Most NL2SQL systems, however, continue to be evaluated on fixed-schema benchmarks, even though the same data is frequently represented using different schemas depending on specific requirements in real-world scenarios. There has been limited research examining how schema design choices influence NL2SQL performance. Accordingly, this work aims to fill this gap by providing a systematic evaluation of the impact of normalization on NL2SQL systems.

\section{Conclusion}
This paper presents the first comprehensive analysis of how normalization influences NL2SQL performance. We found that denormalized schemas, particularly flat ones, perform better for retrieval queries, while normalized schemas are advantageous for aggregation tasks that require handling data consistency issues. Overall, our findings show that the optimal schema for NL2SQL depends on query types and workloads. By bridging database design and NL2SQL, this work highlights the need to explicitly consider schema design for building practical natural language database interfaces.



\bibliographystyle{ACM-Reference-Format}
\balance
\bibliography{citations}

\end{document}